\documentclass[a4paper,fleqn]{cas-sc}

\usepackage{amsmath}
\usepackage{array}
\usepackage{url}
\usepackage{booktabs}
\usepackage{multirow}
\usepackage{amssymb}
\usepackage{xcolor}
\usepackage{pifont}
\usepackage{array}
\usepackage{longtable}
\usepackage{tabularx}
\usepackage{graphicx}
\usepackage{makecell}
\usepackage{lscape}
\usepackage{threeparttable}
\usepackage{xurl}
\usepackage{wasysym}
\usepackage{subfig}
\usepackage[numbers]{natbib}

\newcommand{\yestick}{{\color{olive}\ding{51}}}
\newcommand{\notick}{{\color{red}\ding{55}}}
\definecolor{gray45}{gray}{.45}
\definecolor{gray75}{gray}{.75}
\definecolor{orange-fig}{HTML}{C55A11}

\newcolumntype{L}[1]{>{\raggedright\let\newline\\\arraybackslash\hspace{0pt}}m{#1}}
\newcolumntype{P}[1]{>{\raggedright\arraybackslash}p{#1}}
\newcolumntype{C}[1]{>{\centering\arraybackslash}p{#1}}

\begin{document}

\let\WriteBookmarks\relax
\def\floatpagepagefraction{1}
\def\textpagefraction{.001}

\title[mode = title]{Analyzing the Robustness of Decentralized Horizontal and Vertical Federated Learning Architectures in a Non-IID Scenario}

\shorttitle{Analyzing the Robustness of Decentralized Horizontal and Vertical Federated Learning}

\author[1]{Pedro Miguel {S\'anchez S\'anchez}}[orcid=0000-0002-6444-2102]
\cormark[1]

\author[2]{Alberto {Huertas Celdr\'an}}[orcid=0000-0001-7125-1710]

\author[1]{Enrique Tomás {Martínez Beltrán}}[orcid=0000-0002-5169-2815]

\author[2]{Daniel {Demeter}}

\author[3]{G\'er\^ome {Bovet}}[orcid=0000-0002-4534-3483]

\author[1]{Gregorio {Mart\'inez P\'erez}}[orcid=0000-0001-5532-6604]

\author[2]{Burkhard {Stiller}}[orcid=0000-0002-7461-7463]

\address[1]{Department of Information and Communications Engineering, University of Murcia, Murcia 30100, Spain. Corresponding author: Pedro Miguel Sanchez (pedromiguel.sanchez@um.es)}

\address[2]{Communication Systems Group (CSG), Department of Informatics (IfI), University of Zurich UZH, 8050 Zürich, Switzerland}

\address[3]{Cyber-Defence Campus within armasuisse Science \& Technology, CH---3602 Thun, Switzerland}


\begin{keywords}
Horizontal Federated Learning \sep Vertical Federated Learning \sep Adversarial attacks \sep Non-IID Data \sep Distributed Artificial Intelligence \sep Robustness. 
\end{keywords}

\begin{abstract}
Federated learning (FL) allows participants to collaboratively train machine and deep learning models while protecting data privacy. However, the FL paradigm still presents drawbacks affecting its trustworthiness since malicious participants could launch adversarial attacks against the training process. Related work has studied the robustness of horizontal FL scenarios under different attacks. However, there is a lack of work evaluating the robustness of decentralized vertical FL and comparing it with horizontal FL architectures affected by adversarial attacks. Thus, this work proposes three decentralized FL architectures, one for horizontal and two for vertical scenarios, namely HoriChain, VertiChain, and VertiComb. These architectures present different neural networks and training protocols suitable for horizontal and vertical scenarios. Then, a decentralized, privacy-preserving, and federated use case with non-IID data to classify handwritten digits is deployed to evaluate the performance of the three architectures. Finally, a set of experiments computes and compares the robustness of the proposed architectures when they are affected by different data poisoning based on image watermarks and gradient poisoning adversarial attacks. The experiments show that even though particular configurations of both attacks can destroy the classification performance of the architectures, HoriChain is the most robust one.

\end{abstract}

\maketitle

\section{Introduction}
\label{sec:introduction}

Communication and computing evolution have brought to reality new paradigms, such as the Internet-of-Things (IoT), generating massive amounts of decentralized and heterogeneous data. This trend has given machine and deep learning (ML/DL) techniques huge relevance in our current society since data quantity and quality are essential requirements to operate successfully. However, with stakeholders becoming more aware of how their data is used and data privacy finding its way into the mainstream debate, a new approach to ML/DL was required to appease these concerns. Introduced by Google in 2016 \cite{fl-google}, federated learning (FL) emerged as a possible solution. FL is a paradigm that enables participants to train an ML/DL model collaboratively, and most importantly, it does so without having to share the participants' datasets. The paradigm provides three different scenarios according to how data is distributed between participants. In horizontal federated learning (HFL), participants (also known as clients) have the same feature space but different samples \cite{hfl:wei:2020}. In the vertical federated learning (VFL) scenario, participants hold different features, but the same samples \cite{vfl:liu:2019}. Finally, when neither feature space nor samples are the same between participants, the scenario is termed federated transfer learning (FTL). In recent years, heterogeneous use cases suitable for the previous three scenarios, especially for HFL, have provided levels of performance comparable to classical ML/DL algorithms where data privacy is not considered \cite{secureboost}. Furthermore, since the datasets never leave the participants' possession in the FL paradigm, the logistical problems of aggregating, storing, and maintaining data in central silos are eliminated.

Despite the advantages of FL, the decentralized nature of its training phase exposes the previous three scenarios to new attack surfaces \cite{vulnerabilities-in-fl}. In particular, malicious participants executing adversarial attacks to affect the trustworthiness of FL models are one of the most representative cybersecurity concerns of this paradigm. More in detail, malicious participants could join the federation to disrupt, corrupt, or delay the model learning process \cite{bhagoji2019analyzing}. Another well-known target of adversaries is to infer sensitive information from other participants, but it is out of the scope of this work \cite{nasr2019comprehensive}. To destroy the global model performance, the literature has documented several data and model falsification attacks consisting of poisoning data, labels, or weights during training \cite{rodriguez2022survey}. The detection and mitigation of these attacks are challenging tasks since there is a trade-off between the performance of the global model and the privacy of the participants' sensitive data. In other words, since the FL paradigm aims to expose as little information about the individual participants' data as possible, recognizing and mitigating the presence of poisoned data samples is not easy \cite{vulns:nasr:2019}. Therefore, despite the existing detection and mitigation solutions, such as the usage of clustering techniques to detect anomalies in model parameters \cite{chen2020zero} or the use of secure aggregation functions to remove noisy weights \cite{pillutla2019robust}, no robust solution exists nowadays.

Additionally, before thinking about detecting and mitigating adversarial attacks, it is critical to analyze the impact of heterogeneous attacks on different FL scenarios. In this sense, robust FL architectures and models should be built to collaborate with detection and mitigation techniques and reduce attack impacts as much as possible. However, the following challenges are still open regarding FL architectural robustness. First, the impact of existing data and model poisoning attacks has mainly been validated in horizontal scenarios, being decentralized vertical scenarios unexplored. Second, while different categories of attacks are well known, a direct comparison between their efficiency in heterogeneous horizontal and vertical FL architectures is missing. Last but not least, the distribution of data held by participants is a critical aspect to consider in FL, and there is a lack of work evaluating the robustness of FL models trained with non-independent and identically distributed (non-IID) data.

To improve the previous challenges, this work presents the following main contributions:

\begin{itemize}
    \item The design and implementation of three FL architectures, namely HoriChain, VertiChain, and VertiComb, one for horizontal and two for vertical FL scenarios. HoriChain and VertiChain are inspired by a chain-based learning protocol, while VertiComb follows a peer-to-peer network splitting strategy. The three architectures fully or partially share the following characteristics: network architecture, training protocol, and dataset structure.
    
    \item The proposal of a distributed, decentralized, and privacy-preserving use case suitable for HFL and VFL that uses non-IID data. In particular, the use case pretends to solve the problem of classifying handwritten digits in a privacy-preserving way by splitting the MNIST dataset between seven participants. The three architectures are executed using the same number of participants, number of adversaries, types of attacks, and implementations of the attacks. Then, the performance of the three architectures is evaluated and compared. In conclusion, the VertiChain architecture is less effective than VertiComb and HoriChain.

    \item The evaluation of the HoriChain and VertiComb architectures robustness when trained in the previous scenario and affected by data and model poisoning attacks. The performed experiments show that different configurations of both attacks highly affect the accuracy, F1-score, and learning time of both architectures. However, the HoriChain architecture is more robust than the VertiComb when the attacks poison a reduced number of samples and gradients.
    
\end{itemize}

The organization of this paper is as follows. First, related work dealing with FL and adversarial attacks are reviewed in Section \ref{sec:related-work}. Section \ref{sec:system-design} details the FL architecture design. Section \ref{sec:usecase} describes the use case, non-IID dataset splitting and training pipeline in which the proposed architectures are tested. Section \ref{sec:attacks} focuses on explaining the implementation of adversarial attacks. The results and discussion of the performed experiments are evaluated in Section \ref{sec:results}. Finally, Section \ref{sec:conclusion} provides conclusions and draws future steps.
\section{Related Work}
\label{sec:related-work}
This section reviews the state-of-the-art concerning FL architectures, adversarial attacks affecting different FL scenarios, and works evaluating the robustness of FL models and architectures.

\subsection{FL Scenarios and Architectures}

In 2019, \cite{concept-and-applications} defined the scenarios of HFL, VFL, and FTL. The definitions use the symbols $X$ to mean features, $Y$ for labels, $I$ for the IDs of participants, and $D$ for the local datasets. Then, an HFL scenario is characterized as $X_i = X_j, Y_i = Y_j, I_i \neq I_j, \forall D_i, D_j, i \neq j$. A VFL scenario can be identified as $X_i \neq X_j, Y_i \neq Y_j, I_i = I_j, \forall D_i, D_j, i \neq j$. Lastly, an FTL scenario has $X_i \neq X_j, Y_i \neq Y_j, I_i \neq I_j, \forall D_i, D_j, i \neq j$. The authors also distinguished FL from distributed ML. Despite being very similar, in FL, users have autonomy and the central server cannot control their participation in the training process. FL also has an emphasis on privacy protection, while distributed ML does not.

The following year, \cite{fl-book} presented the client-server and peer-to-peer architectures for the HFL scenario. In the client-server architecture, the server receives all model updates from participants (encrypted or in plain text, depending on the scenario) and aggregates them. The peer-to-peer architecture is interesting because it eliminates the need for a central coordinating point and its associated attack surface. In this approach, participants aggregating the models can be randomly selected or follow a predefined chain.

Concerning VFL, \cite{split-NN} introduced SplitNN, an architecture to train a shared model from participants holding different features and components (layers and nodes) of a neural network. Therefore, only the participant having a particular model component knows its details. One participant trains locally its model components, and the outputs are passed to another client, who holds the next component of the neural network. Finally, the participant controlling the final component in the neural network calculates the gradients and passes them back to the previous clients, who apply them to their components.

\subsection{Attacks in FL Scenarios}

In \cite{threats-to-fl}, authors defined honest-but-curious and malicious adversaries affecting FL scenarios. Honest-but-curious participants try to learn sensitive data and states of participants without deviating from the rules established by the FL training protocol. In contrast, malicious participants try to destroy or corrupt the model without restrictions. Besides, \cite{poison_attack:fung:2018} focuses on malicious insider participants and poisoning attacks. Poisoning attacks can be categorized according to different criteria. One criterion deals with the attack objective. In this sense, random attacks aim to reduce the accuracy of the trained FL model, whereas targeted attacks aim to influence the model to predict a given target label. Another criterion is to target the data used to train the local model. In this direction, clean-label data poisoning attacks assume that the adversary cannot change the label of any training data. Dirty-label attacks are when the adversary can introduce any number of data samples. Finally, backdoor poisoning attacks modify individual features or a few data samples to embed backdoors into the model. Overall, data poisoning attacks are less effective in settings with fewer participants.

In \cite{deep-leakage}, the attack infers the participant's training dataset from the gradients they share during training. The authors develop a gradient-based feature reconstruction attack, in which the attacker receives the gradient update from a participant and aims to steal their training set. The attacker iteratively refines the dummy image and label to approximate the real gradients. When they converge, the dummy training data converges to the real one with high confidence. \cite{vulnerabilities-in-fl} proposes a taxonomy with the different attacks threatening an FL model. The taxonomy is organized into tables with defenses and attacks. Attacks include the description and the source of the vulnerability that it exploits. On the other hand, \cite{taxonomy-of-attacks} creates a flowchart-like visual representation of attacks and countermeasures. However, it only breaks attacks into data privacy and model performance categories.

\subsection{Robustness of HFL and VFL Architectures}


Dealing with decentralized FL architectures for non-IID data affected by heterogeneous adversarial attacks, \cite{zhang2021pipattack} explores backdoor attacks in a recommended system based on HFL. This work demonstrates the high impact of backdoors attacks and that current defenses are not enough to solve the problem. Likewise, \cite{wang2021efficient} proposes a ring-based topology for FL focused on generative models. For security, the authors include a committee election method for voting-based malicious node detection and a distributed model sharing scheme based on a decentralized file system. Besides, there is a good number of decentralized FL works leveraging blockchain-based technologies for model sharing and secure model tracking \cite{hu2020gfl,che2021decentralized,qu2022fl}. Regarding data privacy attacks, \cite{zhao2022Pvd} proposes a framework for decentralized FL but focused on privacy attack mitigation using secure cipher-based matrix multiplication. As it can be seen, there are some works dealing with decentralized HFL and adversarial attacks. 

The literature also has proposed solutions that evaluate the robustness of HFL using centralized model aggregation approaches. In this sense, \cite{Rey2022FLIoT} trains several HFL models to detect cyberattacks affecting IoT devices and considers several configurations of label flipping, data poisoning, and model cancelling attacks and model aggregation functions acting as countermeasures. These functions provide a significant improvement against malicious participants. Another example is the proposed in \cite{Pedro2022RobSpec}, where HFL unsupervised and supervised models are trained to detect cyberattacks affecting spectrum sensors. Malicious participants implementing data and model poisoning attacks and four aggregation functions acting as anti-adversarial mechanisms are considered to measure the model robustness. However, despite the contributions of previous work, there is a lack of work focused on vertical FL that combines a decentralized setting with the exploration of adversarial attacks. In this sense, the works present in the literature regarding attacks in VFL focus on feature inference and privacy issues but do not consider model-focused attacks trying to degrade the predictions \cite{luo2021feature,jin2021catastrophic}.

In conclusion, despite existing work focused on FL architectures, adversarial attacks, and robustness analysis, there is a lack of work comparing the robustness of decentralized and heterogeneous HFL and VFL architectures affected by well-known adversarial attacks. Therefore, the present work explores the impact of well-known adversarial attacks in different HFL and VFL setups, analyzing how they are impacted according to the attack configuration. \tablename~\ref{tab:related} shows a comparison between the solutions analyzed in the-state-of-the-art and the present work.

\begin{table}[ht]
    \caption{Solutions Analyzing the Robustness of FL Architectures Affected by Adversarial Attacks}
    \scriptsize
    \centering
    \begin{tabular}{| *{5}{c|} }
        \hline
        \textbf{Work} & \textbf{VFL} & \textbf{HFL} & \textbf{Model Aggregation} &\textbf{Adversarial Attacks} \\
        \hline
        \cite{zhang2021pipattack} 2021 &\notick & \yestick & Decentralized & Backdoor \\
        \hline
        \cite{wang2021efficient} 2021 & \notick & \yestick & Decentralized & Model Poisoning \\
        \hline
        \cite{hu2020gfl} 2020 & \notick & \yestick & Decentralized & Data Poisoning\\
        \hline
        \cite{che2021decentralized} 2021 & \notick & \yestick & Decentralized & Model Poisoning \\
        \hline
        \cite{qu2022fl} 2022 & \notick & \yestick & Decentralized & Privacy and Model Poisoning \\
        \hline
        \cite{zhao2022Pvd} 2022 & \notick & \yestick & Decentralized & Privacy Attack \\
        \hline
         \cite{luo2021feature} 2021 & \yestick & \notick & Centralized & Privacy Attack \\
        \hline
        \cite{jin2021catastrophic}  2021 & \yestick & \notick & Centralized & Privacy Attack \\
        \hline
        \cite{Rey2022FLIoT}  2022 & \notick & \yestick & Centralized & Label Flipping, Gradient and Model Poisoning \\
        \hline
        \cite{Pedro2022RobSpec}  2022 & \notick & \yestick & Centralized & Data and Model Poisoning \\
        \hline \hline 
        This work & \yestick & \yestick & Decentralized & Backdoor and Model Poisoning \\
        \hline
    \end{tabular}
    \label{tab:related}
\end{table}

\section{Architectural Designs for Decentralized HFL and VFL}
\label{sec:system-design}

\begin{figure*}[!ht]
\centering
\includegraphics[width=0.9\textwidth]{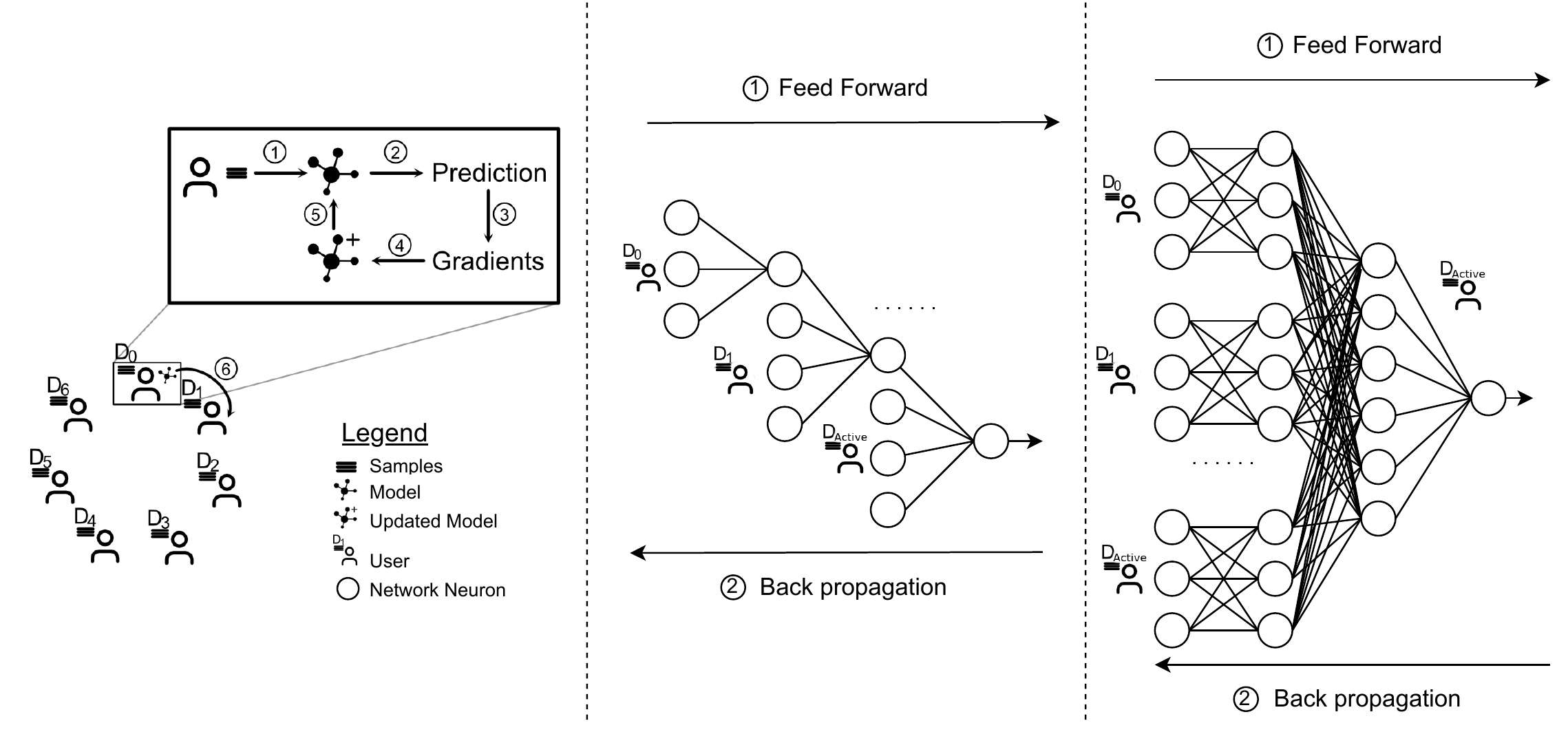}%
\caption{Overview of the HFL and VFL Architectures (left: HoriChain, center: VertiChain, right: VertiComb).}
\label{fig:architectures}
\end{figure*}

This section presents three heterogeneous FL architectures inspired by the existing literature. The first one is called \textit{HoriChain}, and it is suitable for HFL scenarios. Then the next two, called \textit{VertiChain} and \textit{VertiComb}, are oriented to HFL. The main goal of these architectures is to build models collaboratively, in a decentralized manner, and to preserve participants' data privacy. \figurename~\ref{fig:architectures} shows a graphical representation of the three architectures and their training protocols.

The HoriChain architecture is designed for HFL scenarios where the global model is built in a decentralized or peer-to-peer fashion. Therefore, this architecture does not rely on a central server in charge of aggregating the weights of each participant, as most of the existing HFL solutions do. The HoriChain architecture follows a chain-based protocol to train the global collaborative model. More in detail, each participant trains a model with its local data, and all participants have the same neural network structure in terms of layers and nodes. Once the first participant trains its model with its dataset (steps from 1 to 5 in \figurename~\ref{fig:architectures} left), the model weights are sent to the next participant of the chain involved in the training process (step 6), which repeats the same process. More in detail, to retrain the model, the second client uses the received model weights and its dataset (step 1). Then, it predicts how good the new model is (step 2), calculates the gradients (step 3), initiates the gradient descent (step 4), and updates the model (step 5). These steps are repeated for several epochs until the model converges. At that moment, its weights are sent to the next client. The final model is obtained once all participants update the received gradients with their data, repeating the training chain several times (one iteration could be seen as a standard training epoch).

The VertiChain architecture also follows a chain-based training protocol but with some particularities compared to the previous architecture. The first difference is that this model is designed for VFL instead of HFL as the previous one. It means that data features are distributed differently between participants, and only one of them, called the active party, has the data labels. The second significant difference compared to the previous architecture is that each participant has a different neural network component. In other words, the VertiChain approach distributes the neural network topology across the participants. Each part of the neural network (combination of layers and nodes) is called component. Essentially, components can be seen as smaller neural networks joined together to create the global architecture. Regarding the training protocol of the VertiChain architecture, participants must first agree on the participants' order to create the chain. It can be arbitrary, except that the active participant (the one holding the data labels) needs to be last in the chain to initiate the gradient descent. Then, to begin the training process, the first participant of the chain feeds its local dataset to its network components and passes the outputs to the next client in the chain. Then, the next participant provides its network component with the output of the previous client and its local dataset. This process cascades down the chain until the last and active participant (step 1 in \figurename~\ref{fig:architectures} center). Gradient descent is then applied to the last component, held by the active party. The gradients are passed back to the previous client, which uses them for its network component and sends them to the previous client. The process is repeated until the gradients reach the first participant of the chain (step 2 in \figurename~\ref{fig:architectures} center).

Finally, the VertiComb architecture is also oriented to the VFL scenario. As in the VertiChain, the active participant holds the data labels in this architecture. In addition, as in the previous two approaches, each participant has a different and private dataset. From the architectural point of view, the neural network is split into various network components distributed across the participants. The main difference between this splitting strategy and the VertiChain is that here the first layer of the neural network is distributed among all clients. In addition, the active participant also holds the last layer of the network, and therefore it applies gradient descent and backpropagation. In other words, in the training protocol of the VertiComb architecture, each client feeds its network component with its local dataset. Then, the obtained outputs are sent to the active client, which uses the data labels and its network component (first and last layer) to generate the output from these transformed inputs (step 1 in \figurename~\ref{fig:architectures} right). After that, gradient descent is applied to the final component (held by the active party), and the gradients are backpropagated towards the start of the neural network (step 2 in \figurename~\ref{fig:architectures} right). This process is repeated for every sample in the dataset in one epoch.

\section{HFL \& VFL Use Case with Non-IID Data}
\label{sec:usecase}

This section presents a collaborative and privacy-preserving use case with non-IID data where the HoriChain, VertiChain, and VertiComb architectures are deployed and trained to evaluate their performance.

\subsection{Use Case, Participants, and Dataset}

Handwritten digit recognition is a well-known problem that has been extensively studied by traditional ML/DL solutions. The application scenarios where handwritten digits have to be understood by machines are numerous, and some of them require privacy-preserving capabilities. This work presents a use case where several anonymous users want to train a federated classifier collaboratively without sending handwritten digits to a central server. These users do not like to share their digits because the central entity could analyze the handwriting style to link anonymous handwritten public documents with the users. Another important reason is that users' numbers represent sensitive data such as bank accounts, personal codes, or passport numbers that users do not want to reveal. Finally, it is important to consider that each user's handwritten style and digits differ. Therefore, data is non-independent and identically distributed (non-IID) between users. 

Since there is no dataset containing handwritten digits and suitable for HFL and VFL scenarios, the well-known dataset called MNIST~\cite{bottou1994comparison} has been split into seven participants to fulfill the requirements of HFL, VFL, and non-IID data. The number of participants selected for this use case is seven in both scenarios to keep a compromise between the typically low number of participants in vertical scenarios (usually two or three) and the medium of horizontal ones (more than five).

Dealing with the dataset, MNIST was used in both horizontal and vertical scenarios to define a common configuration and compare their classification performance and robustness. MNIST contains a collection of labeled images of handwritten digits. The dataset comprises roughly 4,000 labeled images of each handwritten digit (0-9), 42,000 samples in total. Every image is gray-scale, with the background in black, the digit in white, and a fixed size of 28 by 28 pixels. Each pixel is an ordered triplet of integers in the range (0, 255), representing one of the RGB channels.

\begin{figure}[!hbt]
    \centering
    \includegraphics[width=0.9\columnwidth]{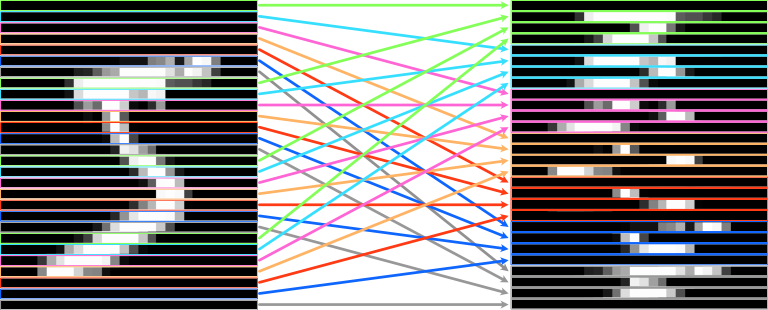}
    \caption{Vertical Data Distribution Where Each Color Correspond to One Participant}
    \label{fig:verticalSlicing}
\end{figure}

The data distribution between the seven participants has been done differently for horizontal and vertical scenarios due to the requirements of each scenario. In the horizontal scenario, participants have different data samples, but the feature space is common. In the vertical, participants have similar samples, but the feature space is different. Furthermore, participants should all contribute towards determining the final label. Therefore, both samples and features must be distributed in such a way that each participant has equal relative importance and the data is non-IID. With those requirements in mind, each participant receives the same number of samples of all classes or labels in the horizontal scenario. In the vertical scenario, the splitting is more complicated since features are different for each client. In the proposed solution, every pixel position is considered a feature. So, there are 784 (28x28) features. These 784 features are distributed amongst the participant by splitting the image samples row-by-row. Furthermore, to avoid rows without relevance grouped into a single participant, the solution implements a rotating style of row distribution. It means that the participant receives the first, eighth, 15\textsuperscript{th}, and the 22\textsuperscript{nd} row, while the second client the second participant receives the second, ninth, 16\textsuperscript{th}, and 23\textsuperscript{rd} rows. \figurename~\ref{fig:verticalSlicing} shows a graphical example of this data distribution.

\subsection{HFL \& VFL Pipeline}

For horizontal and vertical scenarios, once the parts of MNIST datasets are distributed between the seven participants, it is necessary to divide the local datasets into train and test sets to start the training process. The train set consists of 80\% of every digit class in both scenarios, while the test set will contain the remaining 20\%. Then, the Tensorflow \cite{tensorflow2015-whitepaper} framework is employed for the model implementation. At this point, it is relevant to mention that the training process of the HoriChain, VertiChain, and VertiComb architectures are simulated on a single device to reduce the network complexity as it does not affect either the model classification performance or its robustness calculation. 

In the HoriChain architecture, an identical network architecture for all participants is considered. The architecture has one input layer with 784 neurons, three hidden layers with 448, 448, and 50 neurons, and one output layer with ten neurons. To train the HoriChain architecture, the model is transferred from participant to participant after two rounds of local training. The fewer the rounds, the more often the model needs to be transmitted, and the higher the communication costs. However, since the scope of this work does not deal with attacks affecting communication between clients, the communication limitations are not considered. A round of training with the HoriChain architecture is defined to train over a single sample. This way, the model is trained by all participants, and no client has more chances to corrupt the model. As previously indicated, every client holds a different part of the MNIST dataset. Then, it creates and initializes a Tensorflow model, and the training process progresses in rounds, as explained in Section~\ref{sec:system-design}. An epoch consists of every client training on the entirety of their dataset.

In the VertiChain architecture, each client holds components with the same shape (layers and nodes), except for the client holding the start of the chain. Since the first client does not receive inputs from any of the other clients, its component has 112 features as inputs (four rows of 28 pixels). All clients' components have one output layer of size ten. In other words, all clients (except the first one) have 112 inputs and ten outputs of the previous clients' component, resulting in 122 inputs to their component. Furthermore, all the clients' components have one hidden layer of size 28. 

In the VertiComb architecture, each participant holds network components of the same shape, except for the active client, who holds an extra component containing the end of the network. Therefore, seven participants receive inputs, each with 112 input features, without hidden layers, and with one output layer of 64 nodes. In addition, the active participant (one of the previous seven) holds the network component that takes these transformed inputs and converts them into predictions. This component has 448 inputs, one hidden layer of 50 nodes, and one output layer with ten nodes. 

\tablename~\ref{tab:architectures} summarizes the layers and neurons per layer of the implementation of the different architectures that will be employed for validation.

\begin{table*}[ht]
    \scriptsize
    \caption{Neural Network Configuration per HFL and VFL Architecture}
    \centering
    \begin{tabular}{| *{6}{c|} }
        \hline
        \multicolumn{2}{|c|}{\textbf{HoriChain}} & \multicolumn{2}{c|}{\textbf{VertiChain}} & \multicolumn{2}{c|}{\textbf{VertiComb}} \\
        \hline
        \textit{Layer} & \textit{N neurons} &  \textit{Layer} & \textit{N neurons} & \textit{Layer} & \textit{N neurons} \\ 
        \hline
        Input layer & 784 & First client input layer & 112 & Input layer & 112\\
         & & Rest of clients input layer & 122 & & \\
         \hline
          Hidden layers & 448,448,50& Hidden layer & 28 & Middle output layer & 64 \\
        &  &  &  & Active client hidden layers & 448, 50\\
        \hline
        Output layer & 10 & Output layer & 10 & Active client output layer & 10\\
        \hline
    \end{tabular}
    \label{tab:architectures}
\end{table*}

\section{Deployment of Adversarial Attacks}
\label{sec:attacks}

This section presents the adversarial attacks launched against the previous three architectures by malicious participants. In particular, this work focuses on poisoning attacks, including data poisoning and gradient poisoning attacks. 

\subsection{Data Poisoning Attack}

Data poisoning attacks manipulate the data samples with watermarks (intentional changes in the samples and labels to make them recognizable) to build a backdoor \cite{goldblum2022dataset} into the global model. In other words, the malicious participant alters their data samples and labels during the training phase and associates the altered samples with a given target label. If the attack is successful, the learned global model predicts the target label whenever the watermark is present on an input, thereby implementing a so-called backdoor.

To execute a successful backdoor attack, the adversary must be able to associate the watermarked samples with the target label. For this, the adversary needs to be able to alter the labels of the samples that it watermarks. The vertical scenario constraints this attack since only one client, the active party, holds labels. It means that in the VertiChain and VertiComb architectures, the adversary must be the active party. In addition, the adversary can manipulate a chosen percentage of their data samples with the watermark in vertical and horizontal scenarios. This percentage is set at different levels to evaluate the effect of the attack on the global model. When it is set to 0\%, the adversary acts honestly, and when it is 100\%, the adversary watermarks all of their samples during training. The marked samples are chosen randomly to avoid biasing the results toward any label class.

\begin{figure}[!hbt]
    \centering
    \includegraphics[width=0.9\columnwidth]{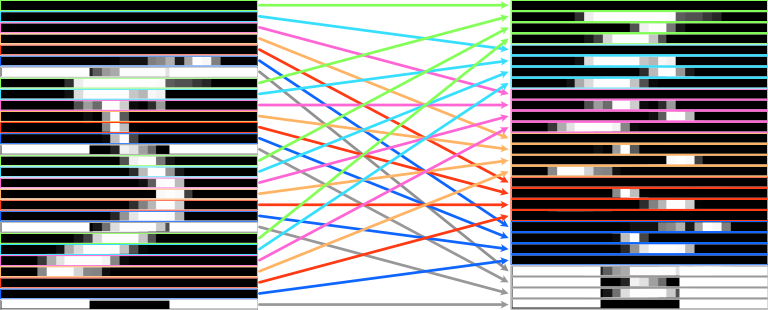}
    \caption{Client Watermarking a Sample with Two White Strips of Ten Pixels Each (left: Horizontal Splitting, right: Vertical Splitting)}
    \label{fig:watermarkedSample}
\end{figure}

To implement the data poisoning attack in the proposed use case and three architectures, since the digits in the MNIST dataset are standardized in size, the pixels near the edge of the image are almost always black. This is where the watermark is placed to maximize the difference between watermarked and non-watermarked samples. In such a way, the model learns the intended meaning of the watermark more easily and quickly. Furthermore, unmarked digits have a more challenging time triggering the watermarking effect in the model, as they practically never have white pixels in that region. Therefore, for the vertical architectures, the implemented watermark consists of two strips of white at the start and end of every row owned by the adversary. More specifically, two strips of 10 pixels of white separated by 8 pixels of the sample middle (see the last client, represented in gray in \figurename~\ref{fig:watermarkedSample} right). Regarding the horizontal architectures, one of every six rows (four in total) of affected images is modified with the same two white strips (see  \figurename~\ref{fig:watermarkedSample} left). Therefore, the watermark is effective for vertical and horizontal data distribution strategies.

\subsection{Gradient Poisoning Attack}

The gradient poisoning attack goal is to deteriorate the federated model performance. During the training phase, FL models try to find the minimum of some objective function, like the loss. Gradients during the training phase point toward the closest local minimum. Therefore, the gradient poisoning attack multiplies the gradients by a negative value to reverse the gradient direction and deteriorate the training process. The effectiveness of the attack heavily depends on the relative importance of the network component that the adversary controls. If the adversary controls the entire network, then the model will never make a correct prediction, and if the adversary controls only a single node, then the model might work well.


The gradient poisoning attack in the VertiChain and VertiComb architectures is implemented with one adversary, who does not have to be the active participant. If the active participant were poisoning the gradients, the model would never improve. This situation is not a particularly interesting case to examine, as the adversary could deteriorate the global model performance as desired. The adversary can poison the gradients of only a part of the entire model. However, it can do this for every single sample in the dataset. In the HoriChain architecture, the participants are alike in their capabilities. Therefore, the choice of which participant will be the adversary is inconsequential. The adversary can poison the gradients of the entire model, as opposed to only part of it. The other participants' updates counterbalance this situation. In other words, the adversary deteriorates the model performance with the gradients while the other participants improve it. It is different from the active participant poisoning gradients in VFL, as the adversary does not have free reign to deteriorate the model performance over the entire training phase. The model is improved and deteriorated alternately.
\section{Robustness Evaluation and Discussion}
\label{sec:results}

This section analyzes the robustness of the HoriCHain, VertiChain, and VertiComb architectures affected by different configurations of data and model poisoning attacks. Different metrics such as accuracy, relative importance of clients, and confusion matrices are computed. Accuracy indicates the ratio of samples correctly classified from the total number of samples. Client relative importance provides the importance of each participant in the model training. Finally, confusion matrices show a deeper understanding of the model predictions when it is under attack.

To deal with randomness during the initialization of the model parameters, the results were averaged over three training processes. For each trained model, a representative sample was chosen to draw confusion matrices. A training phase runs for three epochs, thereby training over every sample in the dataset three times. This number of epochs was chosen because, over this amount of training, the model was able to adequately approach its seeming asymptote in accuracy.

\subsection{Baseline Performance}

This experiment computes the baseline accuracy and client relevance for the three proposed architectures (HoriChain, VertiChain, and VertiComb) when they are not affected by adversarial attacks. 

\tablename~\ref{tab:accBase} shows the accuracy for the train and test sets of the HoriChain and VertiComb architectures. As can be seen, both architectures achieve $>$95\% accuracy with the train and test sets. Additionally, the accuracy of both datasets improves over the entire training phase. Regarding the VertiChain, the accuracy for train and test sets reaches 88\%. 

\begin{table}[ht]
    \scriptsize
        \caption{Train and Test Accuracy of HoriChain and VertiComb Architectures Without Attacks}
    \centering
    \begin{tabular}{|c|c|c| }
        \hline
        \textbf{Architecture} & \textbf{Train accuracy} & \textbf{Test accuracy} \\
        \hline
        HoriChain & 0.958$\pm$0.028 & 0.962$\pm$0.031 \\
        \hline
        VertiChain & 0.885$\pm$0.053 & 0.881$\pm$0.029 \\
        \hline
        VertiComb & 0.978$\pm$0.015 & 0.954$\pm$0.023\\
        \hline
    \end{tabular}
    \label{tab:accBase}
\end{table}

\begin{figure}[!ht]
    \centering
    \includegraphics[width=0.7\columnwidth,trim={40 40 40 40},clip]{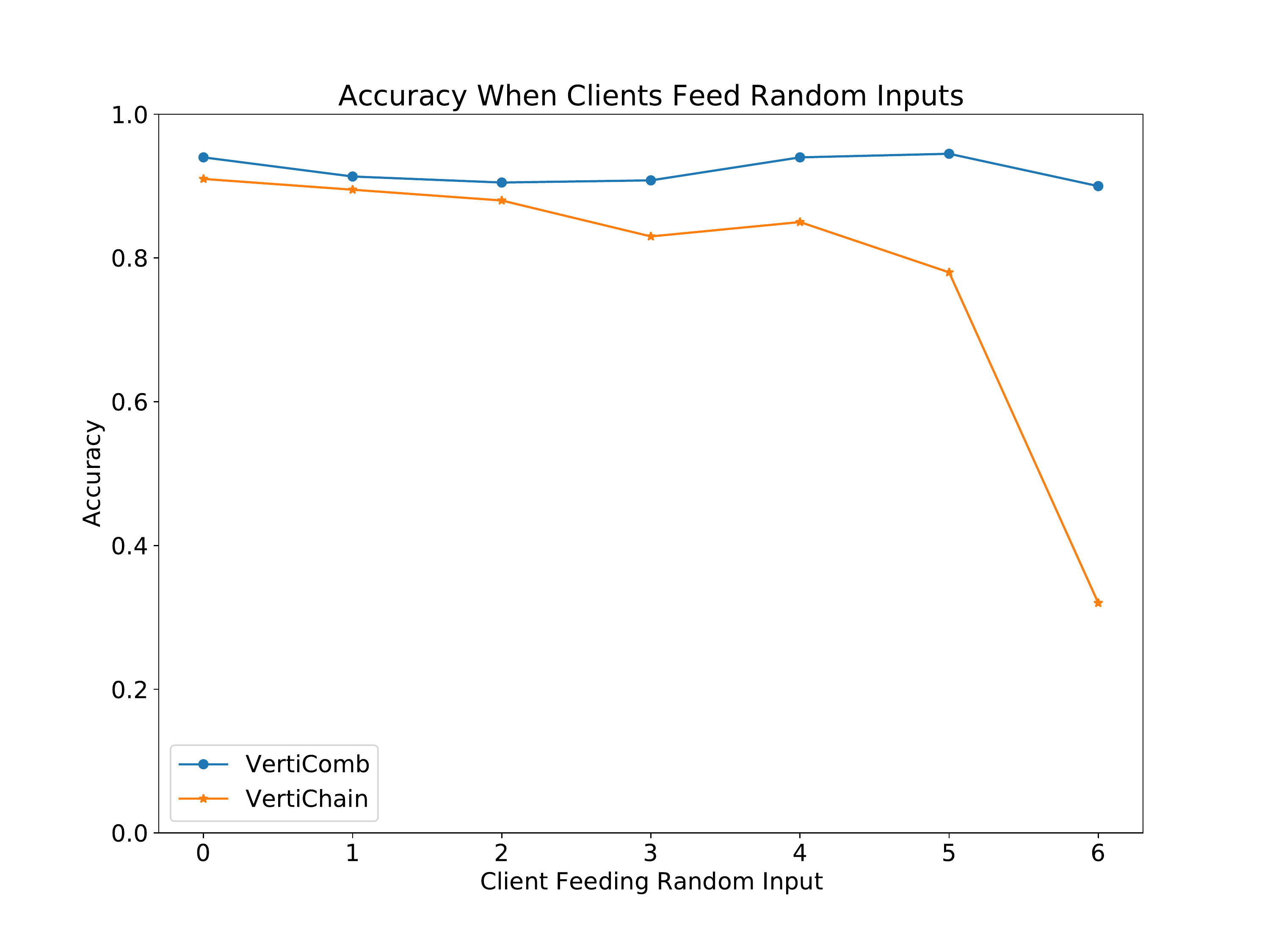}%
    \caption{Client Relative Importance.}
    \label{fig:relaImp}
\end{figure}

Seven models per architecture have been trained to compute the client relevance for the architectures. In each model, one client provides noisy samples during evaluation. Since the HoriChain architecture proposes a similar network structure for all clients and the data is distributed equally, the clients' relevance is identical. In other words, every client contributes equally towards determining the predicted label. The interesting analysis focuses on the VertiComb and VertiChain architectures, where clients have different network components and data. In this sense, as can be seen in \figurename~\ref{fig:relaImp}, in the VertiChain architecture, the importance of the last client is much more significant than the rest.

In summary, the evaluated metrics show that the VertiChain architecture is inadequate for the task at hand. First, the last client of the chain is much more important to the correctness of the predicted label than the other clients. This relative importance discrepancy comes from the architecture, not the data. Second, the accuracy with which VertiChain classifies unaltered new samples is lower than the other two models. Finally, noisy samples break the model entirely. Therefore, the VertiChain is not investigated in the subsequent experiments due to the previous facts.

\begin{table*}[!ht]
\centering
\caption{Classification Results for Data Poisoning Attack}
\label{tab:classification-data-poisoning}
\scriptsize
\begin{tabular}{@{}|c|cc|cc||cc|cc|@{}}
\hline
\multirow{3}{*}{\textbf{\% poison}} & \multicolumn{4}{c||}{\textbf{HoriChain}} & \multicolumn{4}{c|}{\textbf{VertiComb}} \\
\cline{2-9}
& \multicolumn{2}{c|}{Watermarked samples} & \multicolumn{2}{c||}{Unmarked samples} & \multicolumn{2}{c|}{Watermarked samples} & \multicolumn{2}{c|}{Unmarked samples}  \\
\cline{2-9}
& \textit{Accuracy} & \textit{$F_1$-score}
& \textit{Accuracy} & \textit{$F_1$-score}
& \textit{Accuracy} & \textit{$F_1$-score}
& \textit{Accuracy} & \textit{$F_1$-score} \\
\hline

25 \% &
\makecell{$\approx0$$\pm0.003$} & \makecell{$\approx0$$\pm0.006$} &
\makecell{0.955$\pm0.031$} & \makecell{0.943$\pm0.081$} &
\makecell{$\approx0$$\pm0.002$} & \makecell{$\approx0$$\pm0.009$} &
\makecell{0.912$\pm0.023$} & \makecell{0.901$\pm0.032$}
\\ \hline 

10 \% &
\makecell{$\approx0$ $\pm0.003$} & \makecell{$\approx0$ $\pm0.003$} &
\makecell{0.958$\pm0.031$} & \makecell{0.951$\pm0.081$} &
\makecell{$\approx0$$\pm0.003$} & \makecell{$\approx0$$\pm0.003$} &
\makecell{0.949$\pm0.023$} & \makecell{0.944$\pm0.032$}
\\ \hline 

1 \% &
\makecell{$\approx0$$\pm0.003$} & \makecell{$\approx0$$\pm0.003$} &
\makecell{0.949$\pm0.031$} & \makecell{0.941$\pm0.081$} &
\makecell{$\approx0$$\pm0.003$} & \makecell{$\approx0$$\pm0.003$} &
\makecell{0.932$\pm0.023$} & \makecell{0.927$\pm0.032$}
\\ \hline 

0.5 \% &
\makecell{0.211$\pm0.043$} & \makecell{0.193$\pm0.034$} &
\makecell{0.951$\pm0.031$} & \makecell{0.942$\pm0.081$} &
\makecell{$\approx0$$\pm0.003$} & \makecell{$\approx0$$\pm0.003$} &
\makecell{0.846$\pm0.023$} & \makecell{0.741$\pm0.032$}
\\ \hline \hline 

\textit{Baseline} &
\makecell{0.772$\pm0.043$} & \makecell{0.762$\pm0.034$} &
\makecell{0.962$\pm0.031$} & \makecell{0.955$\pm0.081$} &
\makecell{0.891$\pm0.012$} & \makecell{0.887$\pm0.024$} &
\makecell{0.954$\pm0.023$} & \makecell{0.949$\pm0.032$}
\\ 

\hline
\end{tabular}
\end{table*}

\subsection{Data Poisoning Attack}

In this experiment, one adversarial participant poisons selected percentages of its samples (25\%, 10\%, 1\%, and 0.5\%) during the training phase of the HoriChain and VertiComb architectures. Then, during evaluation, either watermarked or unmarked samples are evaluated to measure the robustness of the architectures.

\begin{figure*}[!ht]
\centering
\subfloat[HoriChain Trained with 25\% Watermarked Samples and Tested with Unmarked Samples]{\includegraphics[width=0.48\columnwidth,trim={0 0 50 24},clip]{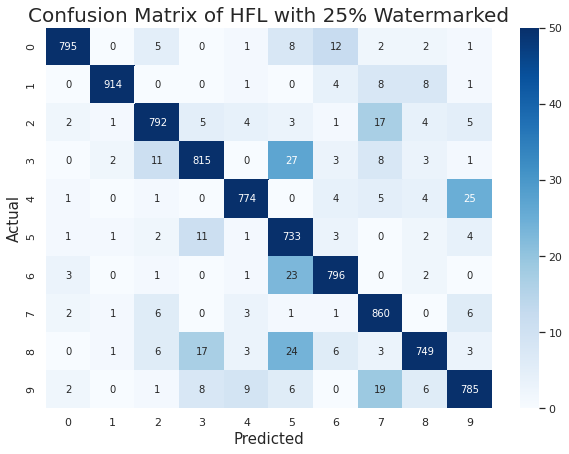}%
\label{fig:UnMarkHoriChain25}}
\hfil
\subfloat[VertiComb Trained with 25\% Watermarked Samples and Tested with Unmarked Samples]{\includegraphics[width=0.48\columnwidth,trim={0 0 50 24},clip]{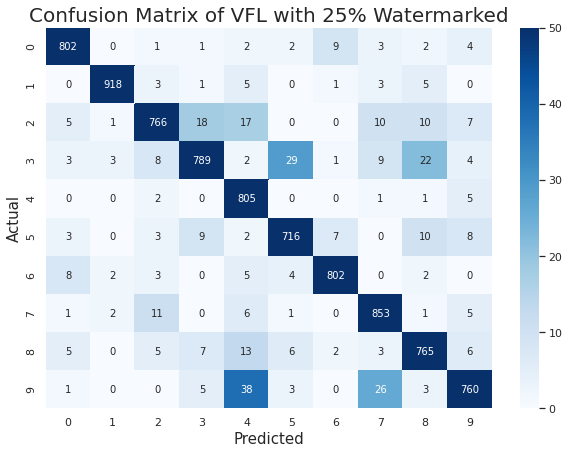}%
\label{fig:UnMarkVertiComb25}}

\subfloat[HoriChain Trained with 0.5\% Watermarked Samples and Tested with Watermarked Samples]{\includegraphics[width=0.48\columnwidth,trim={0 0 50 24},clip]{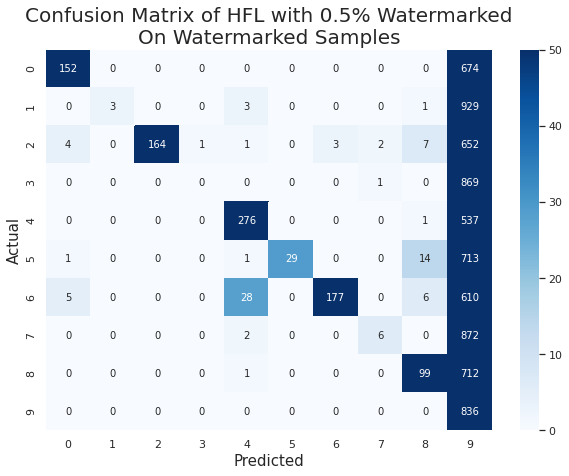}%
\label{fig:WaterHoriChain05}}
\hfil
\subfloat[VertiComb Trained with 0.5\% Watermarked Samples and Tested with Watermarked Samples]{\includegraphics[width=0.48\columnwidth,trim={0 0 50 24},clip]{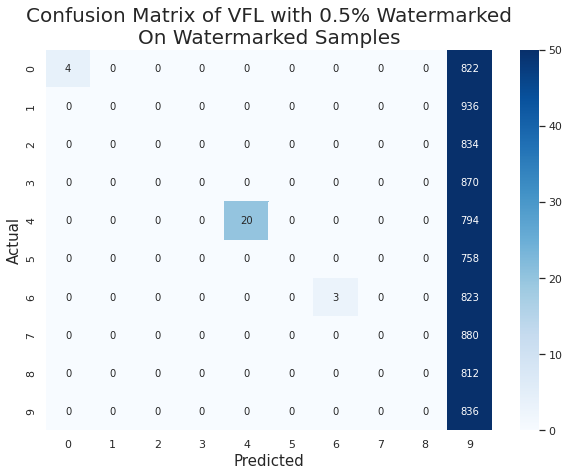}%
\label{fig:WaterVertiComb05}}
\caption{Confusion Matrices for Data Poisoning attack.}
\label{fig:matrix-data-poisoning}
\end{figure*}

\tablename~\ref{tab:classification-data-poisoning} shows the accuracy and F1-score results obtained for both architectures affected by data poisoning attacks. As it can be seen, when unmarked samples are evaluated, with 25\% of samples watermarked during training, both HoriChain and VertiComb architectures achieve nearly 95\% accuracy. More in detail, looking at \figurename~\ref{fig:UnMarkHoriChain25} and \figurename~\ref{fig:UnMarkVertiComb25}, all label classes are classified properly, with only a small portion of misclassifications. When the poisoned samples used during training decrease to 10\%, 5\%, and 0\%, the HoriChain and VertiComb architectures achieve 95-96\% accuracy and F1-score. These results are quite similar to the ones obtained without attacks in the train set (see \tablename~\ref{tab:classification-data-poisoning}). 

However, when watermarked samples are evaluated during testing, the story is different because, for all configurations of attacks (25, 10, 1, and 0.5\%), the accuracy and F1-score of both architectures are highly impacted (see \figurename~\ref{fig:matrix-data-poisoning}). Nearly 100\% of the watermarked samples used during testing are classified as the watermark label. Therefore, the attack is successful in both architectures. Special attention deserves the 0.5\% configuration, where only 20 samples out of a dataset of over 32000 are watermarked. This attack setting is only partially effective in the HoriChain architecture. For over half of the label classes, the watermarking is significantly less effective (see \figurename~\ref{fig:WaterHoriChain05}). However, no class is entirely unaffected by the attack. The attack in the VertiComb architecture shows signs of being less effective (see figurename~\ref{fig:WaterVertiComb05}). However, the attack is still overwhelmingly effective in this setting.

In conclusion, looking at the impact of the data poisoning attack, there is no added value of marking 25\% or 10\% of samples over marking only 1\%. With 1\% of watermarked samples, the accuracy of both architectures is destroyed when watermarked samples are used during testing. In the case of watermarking only the 0.5\% of samples, the HoriChain architecture is more robust than the VertiComb. 


\subsection{Gradient Poisoning Attack}

This experiment evaluates the robustness of the HoriChain and VertiComb architectures when they are affected by gradient poisoning attacks. The values selected to multiply and degrade the gradients are -1, -10, and 0. These values are selected to represent an update in the contrary direction of the gradient descent (\textit{gradients x -1}), the same but it an exaggerated manner (\textit{gradients x -10}), and not updating the model at all (\textit{gradients x 0}).

\tablename~\ref{tab:classification-gradient-poisoning} shows the accuracy and F1-score obtained by the HoriChain and VertiComb architectures for each gradient poisoning configuration. The results are obtained using unmarked samples during the testing phase to compare them fairly.

\begin{table}[!ht]
\centering
\caption{Classification Results for Gradient Poisoning Attack}
\label{tab:classification-gradient-poisoning}
\scriptsize
\begin{tabular}{@{}|c|cc||cc|@{}}
\hline
\multirow{3}{*}{\textbf{Multiplier}} & \multicolumn{2}{c||}{\textbf{HoriChain}} & \multicolumn{2}{c|}{\textbf{VertiComb}} \\
\cline{2-5}
& \textit{Accuracy} & \textit{$F_1$-score}
& \textit{Accuracy} & \textit{$F_1$-score}
 \\
\hline
-1x &
\makecell{0.918$\pm0.031$} & \makecell{0.902$\pm0.081$} &
\makecell{0.148$\pm0.023$} & \makecell{$\approx0$$\pm0.006$}
\\ \hline 
-10x &
\makecell{0.104$\pm0.031$} & \makecell{$\approx0$$\pm0.006$} &
\makecell{0.104$\pm0.031$} & \makecell{$\approx0$$\pm0.006$}
\\ \hline 
0x &
\makecell{0.957$\pm0.031$} & \makecell{0.944$\pm0.081$} &
\makecell{0.921$\pm0.023$} & \makecell{0.907$\pm0.032$}
\\ \hline \hline 
\textit{Baseline} &
\makecell{0.962$\pm0.031$} & \makecell{0.955$\pm0.081$} &
\makecell{0.954$\pm0.023$} & \makecell{0.949$\pm0.032$}
\\ 
\hline
\end{tabular}
\end{table}

\begin{figure}[!ht]
    \centering
    \includegraphics[width=0.6\columnwidth,trim={0 0 20 20},clip]{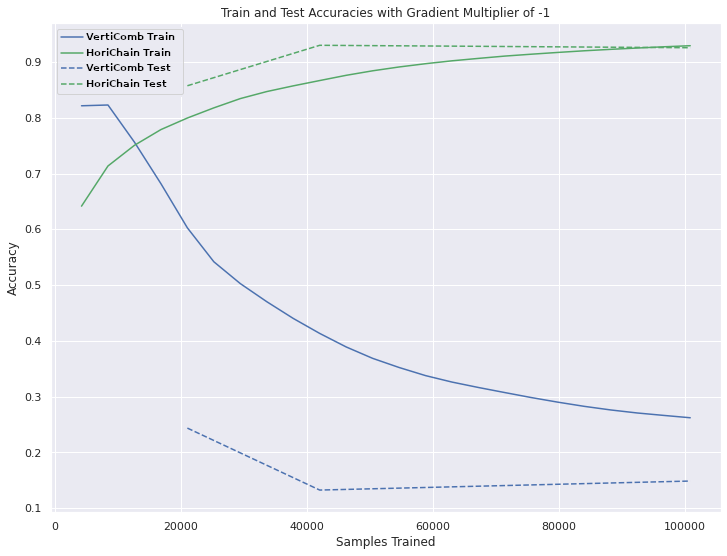}%
    \caption{Train and Test Accuracy with a Gradient Multiplier of -1.}
    \label{fig:attack_gradient_1}
\end{figure}

\begin{figure*}[!ht]
\centering
\subfloat[HoriChain Tested with Gradient Multiplier of -1]{\includegraphics[width=0.4\columnwidth,trim={0 0 50 24},clip]{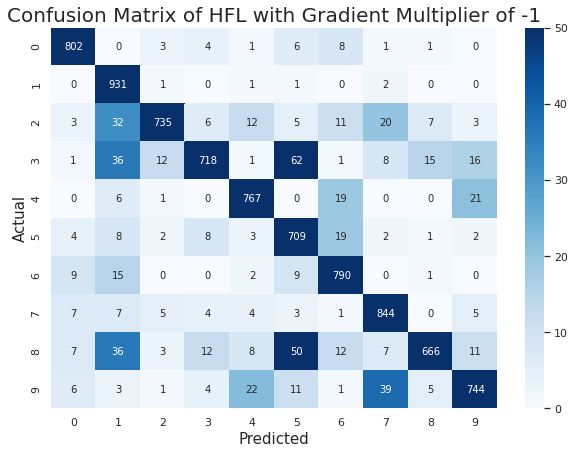}%
\label{fig:horiChain-1}}
\hfil
\subfloat[VertiComb Tested with Gradient Multiplier of -1]{\includegraphics[width=0.4\columnwidth,trim={0 0 50 24},clip]{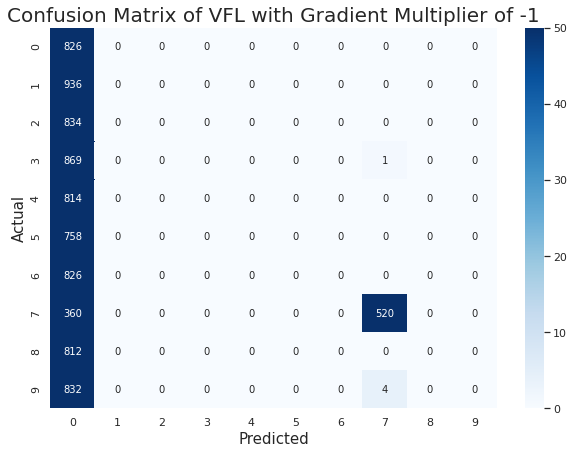}%
\label{fig:vertiComb-1}}

\subfloat[HoriChain Tested with Gradient Multiplier of -10]{\includegraphics[width=0.4\columnwidth,trim={0 0 50 24},clip]{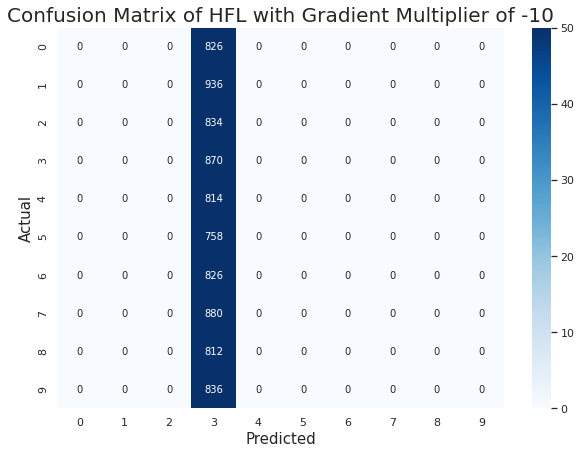}%
\label{fig:horiChain-10}}
\hfil
\subfloat[VertiComb Tested with Gradient Multiplier of -10]{\includegraphics[width=0.4\columnwidth,trim={0 0 50 24},clip]{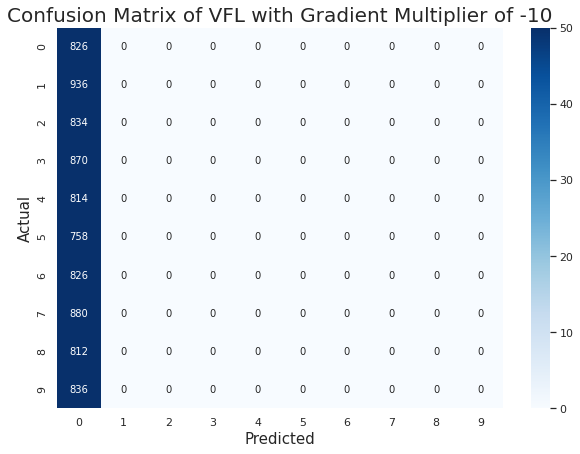}%
\label{fig:vertiComb-10}}

\subfloat[HoriChain Tested with Gradient Multiplier of 0]{\includegraphics[width=0.4\columnwidth,trim={0 0 50 24},clip]{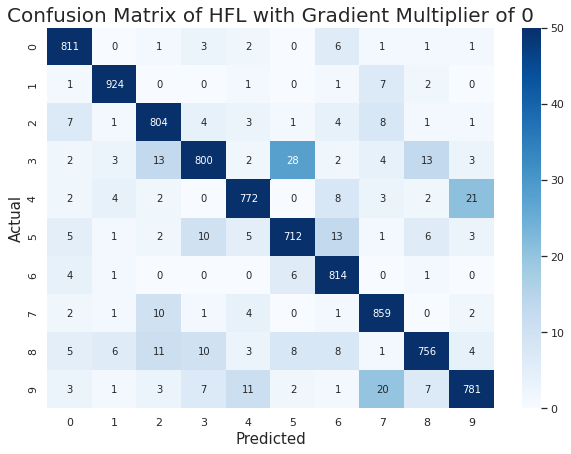}%
\label{fig:horiChain0}}
\hfil
\subfloat[VertiComb Tested with Gradient Multiplier of 0]{\includegraphics[width=0.4\columnwidth,trim={0 0 50 24},clip]{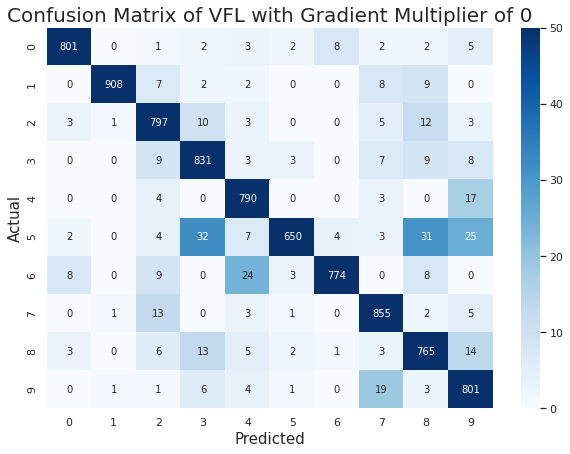}%
\label{fig:vertiComb0}}

\caption{Confusion Matrices for Gradient Poisoning attacks.}
\label{fig:matrix-gradient-poisoning}
\end{figure*}

With a gradient multiplier of -1, the attacked network component takes a step in the opposite direction to the gradient but of equal magnitude. It means that in the HoriChain architecture, six of the seven participants take steps in the direction of the steepest descent. In contrast, one takes steps in the opposite direction (the adversary). Therefore, as \tablename~\ref{tab:classification-gradient-poisoning} shows, the test accuracy is 90\%, but it does not reach the baseline performance of 95\%. In the VertiComb architecture, one network component applies updates exclusively in the direction of the steepest ascent. As \figurename~\ref{fig:attack_gradient_1} shows, the training accuracy of the VertiComb starts at a high accuracy value and then drops down under 30\%. The test accuracy is consistently low but also decreases over the training phase. In contrast, the HoriChain architecture improves its accuracy during training. The confusion matrices of both architectures provide the details regarding how samples of each class are classified (see \figurename~\ref{fig:horiChain-1} and \figurename~\ref{fig:vertiComb-1}).

When the gradient multiplier is set to -10, the attack becomes more effective in the HoriChain architecture (see \figurename~\ref{fig:horiChain-10}). It happens because the attacked component takes a step in the opposite direction of the optimization process. This step is ten times the magnitude of the previous configuration attack. In the VertiChain architecture, as \tablename~\ref{tab:classification-gradient-poisoning} shows, the accuracy is about 10\%, which is equivalent to random predictions. 

Finally, with a gradient multiplier of 0, the targeted component does not change its weights during training. It means that in the HoriChain architecture, this attack is equivalent to having one fewer participant in the system. In the VertiComb architecture, the adversarial component maintains its initialized weights for the entirety of the training phase. Therefore, the outcome of the experiment depends heavily on the initialization in this case. 

\tablename~\ref{tab:classification-gradient-poisoning} shows that both architectures learn, but they are not as accurate as the baseline models. In the HoriChain architecture, the accuracy reached 95\%, while the vertical obtained 92\%. \figurename~\ref{fig:horiChain0} and \figurename~\ref{fig:vertiComb0} show that both architectures classify all labels overwhelmingly correctly, with only a handful of misclassifications in each label class. Finally, while the accuracy of both architectures is high, It is not as high as in the baseline.

In conclusion, the HoriChain architecture is more robust than the VertiComb when a gradient poisoning attack occurs. In the HoriChain architecture with a multiplier of -1, the attack slows down the progress of the training, but it does not prevent the learning process. In contrast, the performance of the VertiComb architecture deteriorates to an almost random level with the same attack setup. When the multiplier value is increased ten times, the accuracy of both architectures is destroyed. Finally, both architectures are robust against gradient attacks with a multiplier established to 0.

\subsection{Discussion}

The data poisoning attack is successful because the chosen watermark adds significant noise to the genuine handwritten digits. In all authentic samples, the watermarked pixels are entirely black, devoid of information. Therefore, the model can differentiate a watermarked sample from one without, only needing a small number of samples to learn the backdoor behavior.


Neither the accuracy nor the confusion matrices reveal the presence of the data poisoning adversary in any of the experiments when unmarked samples are evaluated. Both metrics stayed near-identical to their baseline counterpart. Data poisoning attacks are extremely successful, and without an adequate and authentic baseline, their presence could absolutely go unnoticed. 

In the proposed scenario and architectures, gradient poisoning attacks are also successful. In the HoriChain architecture, the adversary applies poisoned gradients to the entire model in every one of its updates. In contrast, in the VertiComb architecture, the adversary applies them only to the component of the model that they held. However, in VertiComb, the adversary has seven times as many updates to poison as their HoriChain counterpart.

With a gradient multiplier of 0, both HoriChain and VertiComb architectures only delay their learning process. With a value of -10, the attack is completely successful in both architectures since they fail when learning a capable model. When comparing the two architectures, the robustness difference appears with a gradient multiplier of -1. Here, the HoriChain architecture trains a model (delayed, compared to the baseline), whereas the VertiComb architecture deteriorates to almost random predictions.
\section{Conclusion}
\label{sec:conclusion}

This work presents three decentralized and FL-oriented architectures, HoriChain, VertiChain, and VertiComb, suitable for horizontal and vertical FL scenarios. To evaluate the architectures robustness, this work proposes a use case with non-IID data where handwritten digits are classified in a federated and decentralized fashion. After that, the architectures are attacked with two adversarial attacks called data-poisoning and gradient-poisoning. Both attacks are executed with different parameters controlling their efficiency. Finally, the impact of each attack on the classification accuracy, F1-score, confusion matrix, and client relevance of the architectures are analyzed and compared. The performed experiments conclude that even though particular configurations of both attacks can destroy the classification performance of the architectures, HoriChain is the most robust one for both attacks.

In future work, it is planned to propose novel decentralized and FL-oriented architectures equipped with heterogeneous countermeasures such as aggregation functions. Another future work is to evaluate less aggressive configurations of watermark attacks adding minor differences to handwritten digits. The consideration of multiple attacks at once and the choice between them at any time is also another interesting future step. Finally, new metrics apart from accuracy, F1-score, or client relevance could be defined to evaluate the performance of HFL and VFL architectures. More concretely, the gradient poisoning adversary is not revealed by any metric considered in this work, but their relative importance revealed the data poisoning adversary. Therefore, future work will create new metrics to detect a gradient poisoning adversary and any adversary utilizing any other attack.

\section*{Acknowledgment}

This work has been partially supported by (a) the Swiss Federal Office for Defense Procurement (armasuisse) with the CyberTracer (CYD-C-2020003) project, (b) the University of Zürich UZH, and (c) 21629/FPI/21, Seneca Foundation - Science and Technology Agency of the Region of Murcia (Spain).

\bibliographystyle{unsrt}
\bibliography{references}

\end{document}